\documentclass[a4paper,fleqn]{cas-dc}

\usepackage[numbers]{natbib}
\usepackage{tabularx}
\usepackage{bm}
\usepackage{xcolor}

\definecolor{lightblue}{RGB}{45,110,180}

\hypersetup{
	colorlinks=true,
	allcolors=lightblue
}

\def\tsc#1{\csdef{#1}{\textsc{\lowercase{#1}}\xspace}}
\tsc{WGM}
\tsc{QE}
\tsc{EP}
\tsc{PMS}
\tsc{BEC}
\tsc{DE}

\begin{document}
\let\WriteBookmarks\relax
\def\floatpagepagefraction{1}
\def\textpagefraction{.001}
\shorttitle{Kinesthetic teaching via dynamic force guidance}
\shortauthors{C. Li et~al.}

\title [mode = title]{A Unified Dynamic Force Guidance Framework for Performance-Optimized Kinesthetic Teaching}                      
%

\author[1]{Chunxin Li}
\ead{lcx2022@sjtu.edu.cn}

\affiliation[1]{organization={State Key Laboratory of Mechanical System and Vibration, 
		Institute of Robotics, 
		School of Mechanical Engineering, 
		Shanghai Jiao Tong University},
                city={Shanghai},
                postcode={200240}, 
                country={China}}

\author[1]{Jianhua Wu}
\cormark[1]
\ead{wujh@sjtu.edu.cn}

\author[1]{Zhenhua Xiong}
\ead{mexiong@sjtu.edu.cn}

\author[1]{Xiangyang Zhu}
\ead{mexyzhu@sjtu.edu.cn}

\cortext[cor1]{Corresponding author}

\begin{abstract}
Collaborative robots are increasingly deployed in industrial scenarios characterized by frequent product changeovers. As an intuitive programming method, kinesthetic teaching facilitates rapid robot deployment. However, users may overlook the configuration of the robot during kinesthetic teaching, leading to degradation in operational performance. Operational performance refers to the capability of the robot to generate motion and can be quantified by the Minimum Singular Value of the Jacobian matrix. To address this issue, this paper proposes an online dynamic force guidance method that integrates performance constraint and optimization mechanisms. Specifically, variable admittance control maintains the operational performance of the robot above a predefined threshold, while a virtual force actively guides the user to drag the robot towards configurations with improved performance. Experiments are conducted on a 6-DOF collaborative robot, comparing three typical paths in the task space. To evaluate the quality of the taught trajectories, trajectory playback experiments are conducted to analyze the relationship between the operational performance of the robot and the work efficiency. The results demonstrate that the proposed method effectively enhances the operational performance of the robot and consequently improves the work efficiency, holding significant value for reducing production takt time in industrial deployment.
\end{abstract}



\maketitle

\section{Introduction}

The current manufacturing industry is facing rapid changes and personalized production demands \cite{ref1}. Collaborative robots have been extensively applied in industry due to their flexibility and efficiency \cite{ref2,ref3,ref4}. To adapt to continuously changing environment and task requirements, robots need to be reprogrammed quickly \cite{ref31}. Robot programming is a complex and challenging task that requires users to possess profound professional knowledge \cite{ref6}. Kinesthetic teaching is a widely used programming method where human intention is conveyed through applied force and torque to guide robot motion \cite{ref30}. This method enables users to teach new skills to the robot without professional knowledge, and is characterized by intuitiveness and high efficiency \cite{ref7}.

Recently, some researchers have conducted studies on kinesthetic teaching, among which typical methods include impedance and admittance control \cite{ref8}. Admittance control can transform the operational force applied by the user into corresponding motion commands, thereby enabling the robot to exhibit compliance during motion \cite{han2024variable}. The dynamic model of admittance control in Cartesian space can be represented as a mass-spring-damper system:
\begin{equation}\label{eq1}	
	{{\mathbf{M}}_{\mathrm{d}}}(\ddot{\mathbf{X}}-{{\ddot{\mathbf{X}}}_{0}})+{{\mathbf{C}}_{\mathrm{d}}}(\dot{\mathbf{X}}-{{\dot{\mathbf{X}}}_{0}})+{{\mathbf{K}}_{\mathrm{d}}}(\mathbf{X}-{{\mathbf{X}}_{0}})={{\mathbf{F}}_{\mathrm{h}}}
\end{equation}
where $\mathbf{M}_{\mathrm{d}}$, $\mathbf{C}_{\mathrm{d}}$ and $\mathbf{K}_{\mathrm{d}}$ are constant positive definite diagonal matrices, representing virtual mass, damping, and stiffness, respectively. $\mathbf{X}$ denotes the position in Cartesian space, and $\mathbf{X}_{0}$ represents the equilibrium position. $\mathbf{F}_{\mathrm{h}}$ represents the operational force applied by the user, which can be measured by the force sensor installed between the end-effector and the robot.

For kinesthetic teaching, the virtual spring $\mathbf{K}_{\mathrm{d}}$ and equilibrium position $\mathbf{X}_{0}$ are set to zero, thus the admittance control model can be expressed in the form of a mass-damper system:
\begin{equation}\label{eq2}	
	{{\mathbf{M}}_{\mathrm{d}}}\ddot{\mathbf{X}}+{{\mathbf{C}}_{\mathrm{d}}}\dot{\mathbf{X}}={{\mathbf{F}}_{\mathrm{h}}}
\end{equation}

The admittance controller can compute the desired end-effector velocity based on the operational force. Through the Jacobian matrix, the end-effector velocity can be mapped to joint velocities. The obtained joint velocities are sent to the robot controller to achieve motion tracking. For a 6-DOF robot, the relationship between the end-effector velocities and the joint velocities is as follows:
\begin{equation}\label{eq3}	
	\dot{\mathbf{X}}=\left[ \begin{matrix}
		\mathbf{v}  \\
		\bm{\omega}   \\
	\end{matrix} \right]=\left[ \begin{matrix}
		{{\mathbf{J}}_{\mathrm{t}}}  \\
		{{\mathbf{J}}_{\mathrm{r}}}  \\
	\end{matrix} \right]\dot{\mathbf{q}}
\end{equation}
where $\mathbf{v}$ and $\bm{\omega}$ represent the translational and rotational velocities of the end-effector, $\dot{\mathbf{q}}$ represents the joint velocities, $\mathbf{J}_{\mathrm{t}}$ and $\mathbf{J}_{\mathrm{r}}$ denote the translational and rotational components of the Jacobian matrix.

\begin{figure}[pos=!t]
	\centering
	\includegraphics[width=0.6\linewidth]{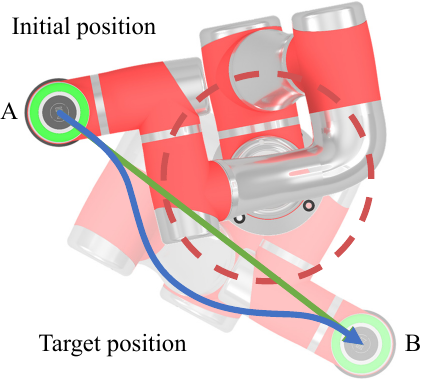}
	\caption{Paths crossing and bypassing the singularity. The red dashed line indicates the shoulder singularity, which is a cylindrical surface centered at the base coordinate origin.}
	\label{fig1}
\end{figure}

Through admittance control, the user can intuitively adjust the position and orientation of the robot. However, since the quality of kinesthetic teaching depends on the operation of the user \cite{ref9}, the operational performance of the robot may degrade during the teaching process. Operational performance refers to the ability of the robot to generate motion in different directions, which is related to the configuration of the robot. As the user is typically unable to perceive the spatial distribution of this performance, the robot may be unintentionally dragged into configurations with low performance \cite{ref32}. This can lead to a decline in the kinematic and dynamic performance of the robot, affecting task accuracy and reliability. A typical example is the kinematic singularity, as shown in \hyperref[fig1]{Fig.~\ref*{fig1}}. When the user drags the robot from point A to point B in a straight path, the robot may pass through the singularity. Singularity can cause the robot to lose certain degrees of freedom, leading to the degradation of the Jacobian matrix \cite{ref10}. At the singularity, the joint velocities calculated through inverse kinematics tend to infinity, leading to system instability and impact loads, which pose significant challenges to the safety of the task \cite{pulloquinga2023admittance}. Although adopting a curved path increases the trajectory length, it can effectively avoid singularities and improve the stability of the system. Therefore, optimizing the operational performance of the robot during the kinesthetic teaching process is of great significance for improving the reliability of the system.

To keep the robot away from singularities, a common approach is to guide the user through the dynamic force. The damped least-squares (DLS) method is a typical approach \cite{ref11}. It obtains feasible joint velocities by minimizing the weighted sum of the tracking error norm and the joint velocity norm. The principle is to add a damping term in the inverse Jacobian expression, thereby ensuring numerical robustness in the neighborhood of singularities \cite{ref33}. However, the DLS method sacrifices the accuracy of the kinematic inverse solution. Based on the DLS method, asymmetric damping applies smaller damping when the robot moves away from the singularity \cite{ref12}. Although damping can alleviate poor conditions near singularities, it cannot actively guide the user to move the robot away from them. Virtual force is another feasible approach. It generates an active force to provide the user with auxiliary feedback and guides the user to drag the robot in a direction that optimizes performance. The direction of the virtual force is usually selected as the gradient of the performance index, and its magnitude depends on the value of the chosen performance index. Some studies have mentioned methods that use the Manipulability Index and the Minimum Singular Value as indices to generate virtual force \cite{ref13,ref14}. However, there are still several problems with using virtual force as a constraint. For example, to ensure that the virtual force can prevent the user from guiding the robot towards a singularity, it is commonly set to an unbounded form, which may lead to stability issues. Besides, using virtual force as a constraint may lead to large forces in uncertain directions, causing the robot to move in a way that does not align with the intention of the user. This can make the robot deviate from user control and reduce the comfort of kinesthetic teaching. In addition, by optimizing the cost function of configuration control to force the inverse matrix to be full rank, the inverse kinematic solution near singularities can be achieved \cite{ref34}.

To construct a reasonable dynamic force generation mechanism, appropriate performance indices must be selected to quantitatively describe the operational performance of the robot. The manipulability ellipsoid can effectively reflect the capability of the robot in all directions. On this basis, performance indices derived from the Jacobian matrix, such as the Manipulability Index \cite{ref15}, the Condition Number (CN) \cite{ref16}, and the Minimum Singular Value (MSV) \cite{ref17}, play important roles in robot control. Since the Jacobian matrix depends on the spatial configuration of the robot, these indices reflect local properties of the robot. The Manipulability Index considers motion in all directions of the end-effector, while the MSV and CN take only certain directions into consideration \cite{ref18}. Directly computing these indices from the Jacobian matrix presents a problem, as the resulting values lack physical consistency \cite{ref19}. For general robots, the Manipulability Index can be decomposed into a translational manipulability measure and a rotational manipulability measure to constrain the range of the overall index \cite{ref20}. Similarly, other indices can be defined in translational and rotational subspaces to evaluate performance \cite{ref21}. However, such approaches still have problems. For example, when considering translational indices, it allows the robot to rotate arbitrarily, while the actual rotation is known. As a result, the computed performance index does not accurately reflect the true operational performance of the robot. Some studies have introduced homogeneous indices that combine translational and rotational performance \cite{ref22}. This paper mainly focuses on the more common case in kinesthetic teaching where the end-effector maintains a constant orientation.

Due to the complex nonlinear relationship between robot performance and its configuration, users find it difficult to perceive and optimize performance intuitively during kinesthetic teaching. Therefore, it is expected that the robot can provide real-time performance feedback through dynamic force during the process, guiding the user to optimize the taught trajectory. This goal requires a quantitative description of robot operational performance, as well as the design of a suitable dynamic force generation mechanism to achieve effective guidance. Furthermore, although several indices have been used to evaluate operational performance, the intrinsic relationship between operational performance and work efficiency should be explored to provide theoretical support for reducing production takt time in industrial deployment.

To address these challenges, an online dynamic force guidance method that integrates performance constraint and optimization mechanisms is proposed. In this method, the performance constraint adjusts the admittance parameters to suppress motion in low-performance directions, while the performance optimization generates a virtual force that actively guides the user towards configurations with better performance. Compared with existing methods, the main contributions of this paper include:
\begin{itemize}
	\item An online dynamic force guidance method based on real-time performance monitoring is proposed. By integrating performance constraint and optimization mechanisms, this method effectively enhances the demonstration quality during kinesthetic teaching.
	\item Variable admittance control suppresses motion towards low-performance directions, while a bounded virtual force provides active guidance without introducing significant interference to the user. Furthermore, a stability analysis based on bounded energy dissipation is provided to prove the stability of the interactive system.
	\item The relationship between operational performance and work efficiency is quantitatively revealed through trajectory playback experiments under joint velocity saturation limits. By recording actual execution time to evaluate the trajectory quality, a theoretical basis is provided for reducing production takt time in industrial deployment.
\end{itemize}

The rest of this paper is organized as follows. Section \ref{sc2} introduces the proposed method and its control framework. Section \ref{sc3} presents the experimental setup and introduces the proposed trajectory quality evaluation metric. Section \ref{sc4} presents and discusses the results. Finally, Section \ref{sc5} provides the conclusion.

\section{Method}\label{sc2}

The proposed method is based on the admittance control framework. It aims to use the dynamic force during kinesthetic teaching to provide the user with auxiliary feedback, thereby guiding the user to avoid low-performance regions and approach configurations with optimized performance. The dynamic force guidance method should include the following characteristics:
\begin{itemize}
	\item The method suppresses user actions that move the robot towards singularities, preventing unstable motion caused by singularity.
	\item The dynamic force is asymmetric, with a strong damping effect when the robot approaches a singularity and a moderate damping effect when moving away from it.
	\item The dynamic force is proactive, capable of actively guiding the user during kinesthetic teaching, and remains bounded in magnitude.
\end{itemize}

\begin{figure*}[pos=!t]
	\centering
	\includegraphics[width=0.7\linewidth]{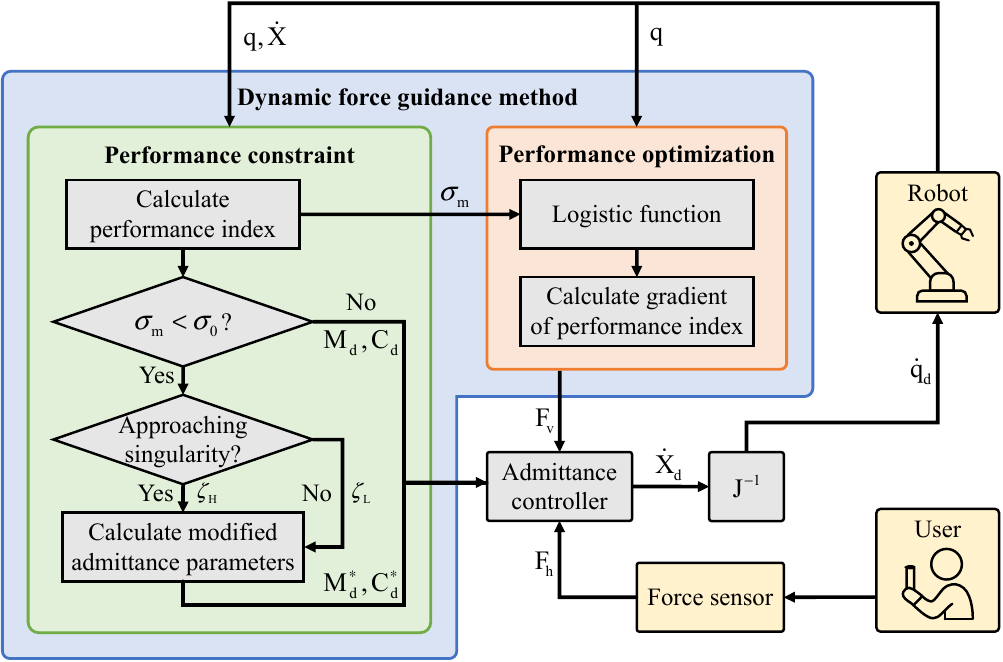}
	\caption{Block diagram of the control structure. The performance constraint module dynamically adjusts the controller parameters based on the robot state. The performance optimization module generates a virtual force, which is combined with the operational force and sent to the adjusted admittance controller. The controller outputs the desired end-effector velocity, which is converted to joint velocity via the inverse Jacobian matrix and sent to the robot.}
	\label{fig2}
\end{figure*}

To achieve these requirements, the dynamic force guidance scheme consists of two main components: performance constraint based on variable admittance control and performance optimization based on virtual force. When the performance index is lower than a set threshold, asymmetric damping is applied to the motion of the robot through variable admittance control. The damping is scaled using a quadratic function, ensuring its continuity. Virtual force is used to actively guide kinesthetic teaching. The magnitude of the virtual force is determined by the performance index, and its direction corresponds to the gradient of the performance index. The magnitude and effective range of the dynamic force can be reasonably adjusted according to the task requirements. A block diagram of the control structure is shown in \hyperref[fig2]{Fig.~\ref*{fig2}}.

\subsection{Selection of Performance Index}

To quantify the operational performance of the robot, the MSV is employed as the performance index. MSV is the Minimum Singular Value of the Jacobian matrix, which characterizes the weakest direction of the motion ability of the robot in Cartesian space \cite{ref24}. By optimizing the MSV, the operational performance of the robot can be improved. Since noncommensurate systems do not have a physically consistent singular value \cite{ref19}, directly computing the MSV from the Jacobian matrix presents problems. Many collaborative tasks place more emphasis on the translational motion of the end-effector \cite{ref25}. Therefore, the motion of the robot end-effector is constrained to translation with a fixed orientation to ensure physical consistency, and this approach can fully reflect the actual operational situation. The translational Jacobian matrix can be expressed as \cite{ref20}:
\begin{equation}\label{eq4}	
	{{\bar{\mathbf{J}}}_{\mathrm{t}}}={{\mathbf{J}}_{\mathrm{t}}}(\mathbf{I}-\mathbf{J}_{\mathrm{r}}^{\dagger }{{\mathbf{J}}_{\mathrm{r}}})
\end{equation}

The singular value decomposition of the translational Jacobian matrix can be expressed as:

\begin{equation}\label{eq5}	
	{{\bar{\mathbf{J}}}_{\mathrm{t}}}={{\mathbf{U}}_{\mathrm{t}}}{{\bm{\Sigma} }_{\mathrm{t}}}\mathbf{V}_{\mathrm{t}}^{T}
\end{equation}
where $\mathbf{U}_{\mathrm{t}}$ and $\mathbf{V}_{\mathrm{t}}$ are orthogonal matrices. The column vectors of $\mathbf{U}_{\mathrm{t}}$ represent the principal axes of the manipulability ellipsoid. The elements on the main diagonal of $\bm{\Sigma}_{\mathrm{t}}$ represent the singular values in each direction, reflecting the operational performance of the robot in different directions, among which the minimum value $\sigma_{\mathrm{m}}$ is the MSV.

\subsection{Performance Constraint Based on Variable Admittance Control}

Through the singular value decomposition of the translational Jacobian matrix, the operational performance of the robot in each principal axis direction can be obtained. In the direction corresponding to the MSV, the robot exhibits the weakest operational performance. When the MSV decreases to a sufficiently low level, dragging the robot along the corresponding direction leads to a degradation in operational performance, causing the joint velocities of the robot to diverge. To prevent the user from continuing to move the robot along the direction of the MSV, the proposed method applies damping to kinesthetic teaching by increasing the admittance parameters. Meanwhile, less damping is applied during kinesthetic teaching when moving away from singularities.

To achieve the above functions, a performance index threshold denoted as $\sigma_0$ is defined. When ${{\sigma }_{\mathrm{m}}}\le {{\sigma }_{0}}$, it indicates that the robot has approached a singularity. Therefore, the performance constraint strategy is activated and the admittance parameters in the direction of the MSV is gradually increased as well. To increase the admittance parameters along the direction of MSV, the original parameter matrix in Cartesian space can first be modified, and then the admittance parameters in the desired direction can be increased through matrix similarity transformation. Considering that the virtual mass $\mathbf{M}_{\mathrm{d}}$ and damping $\mathbf{C}_{\mathrm{d}}$ jointly determine the dynamic characteristics and stability margin of the system \cite{ref35}, $\mathbf{M}_{\mathrm{d}}$ and $\mathbf{C}_{\mathrm{d}}$ are adjusted synchronously, and the original admittance parameters are modified as follows:
\begin{equation}\label{eq6}	
	{{\mathbf{M}}_{\mathrm{tmp}}}=\left[ \begin{matrix}
		m & 0 & 0  \\
		0 & m & 0  \\
		0 & 0 & \mu m  \\
	\end{matrix} \right],{{\mathbf{C}}_{\mathrm{tmp}}}=\left[ \begin{matrix}
		c & 0 & 0  \\
		0 & c & 0  \\
		0 & 0 & \mu c  \\
	\end{matrix} \right]	
\end{equation}
where $\mu$ is a coefficient related to the performance index. The matrix $\mathbf{U}_{\mathrm{t}}$, obtained from the singular value decomposition of the translational Jacobian matrix, can be used for similarity transformation to rotate the base coordinate frame into the principal axis coordinate frame of the manipulability ellipsoid. The transformed mass and damping matrices are given as:
\begin{equation}\label{eq7}	
	\mathbf{M}_{\mathrm{d}}^{*}={{\mathbf{U}}_{\mathrm{t}}}{{\mathbf{M}}_{\mathrm{tmp}}}\mathbf{U}_{t}^{-1},\mathbf{C}_{\mathrm{d}}^{*}={{\mathbf{U}}_{\mathrm{t}}}{{\mathbf{C}}_{\mathrm{tmp}}}\mathbf{U}_{t}^{-1}
\end{equation}

The expression of $\mu$ in \hyperref[eq6]{(\ref{eq6})} is given as:
\begin{equation}\label{eq8}	
	\mu =\left\{ \begin{array}{*{35}{l}}
		1+(\zeta-1) {{\left( \frac{{{\sigma }_{0}}-{{\sigma }_{\mathrm{m}}}}{{{\sigma }_{0}}} \right)}^{2}}, & 0<{{\sigma }_{\mathrm{m}}}\le {{\sigma }_{0}}  \\
		1, & {{\sigma }_{\mathrm{m}}}>{{\sigma }_{0}}  \\
	\end{array} \right.	
\end{equation}
where $\zeta$ is a proportional coefficient. There are two types of values for the coefficient $\zeta$. When the user moves the robot closer to a singularity, a larger value is selected to apply stronger suppression to kinesthetic teaching. When the user moves the robot away from the singularity, a smaller value is selected to realize an asymmetric suppression effect.

The MSV is used to determine whether the robot is moving towards or away from the singularity. When the motion of the robot causes an increase in the MSV, it is considered as moving away from the singularity. Otherwise, it is considered as moving towards it. Assuming a small displacement $\delta {{\mathbf{X}}_{\mathrm{v}}}\in {{\mathbb{R}}^{3}}$ in the direction of the end-effector velocity, the approximate joint position at the next time step can be estimated using the finite difference method as follows:
\begin{equation}\label{eq9}	
	\hat{\mathbf{q}}_{\mathrm{v}}={{\mathbf{q}}_{0}}+\bar{\mathbf{J}}_{\mathrm{t}}^{\dagger }({{\mathbf{q}}_{0}}) \delta {{\mathbf{X}}_{\mathrm{v}}}
\end{equation}
where $\mathbf{q}_{0}$ represents the joint position at the current time step. Based on the joint position at the next time step, the corresponding $\bar{\mathbf{J}}_{\mathrm{t}}(\hat{\mathbf{q}}_{\mathrm{v}})$ and ${{\hat{\sigma }}_{\mathrm{m}}}$ can be calculated. By comparing $\sigma_{\mathrm{m}}$ and ${{\hat{\sigma }}_{\mathrm{m}}}$, it can be determined whether the robot is moving towards or away from the singularity. The selection of the coefficient $\zeta$ is defined as follows:
\begin{equation}\label{eq10}	
	\zeta=\left\{ \begin{array}{*{35}{l}}
		{\zeta_{\mathrm{H}}}, & {{{\hat{\sigma }}}_{\mathrm{m}}}\le {{\sigma }_{\mathrm{m}}}  \\
		{\zeta_{\mathrm{L}}}, & {{{\hat{\sigma }}}_{\mathrm{m}}}>{{\sigma }_{\mathrm{m}}}  \\
	\end{array} \right.	
\end{equation}

\subsection{Performance Optimization Based on Virtual Force}

Applying performance constraint can effectively prevent the user from moving the robot towards a singularity. However, due to the passive nature of this method, the robot cannot provide active feedback to guide the user in moving towards configurations with optimized performance. Therefore, a virtual force is added on top of the operational force measured by the force sensor to provide active guidance. The virtual force is generated based on the performance index.

\begin{figure}[pos=!t]
	\centering
	\includegraphics[width=0.8\linewidth]{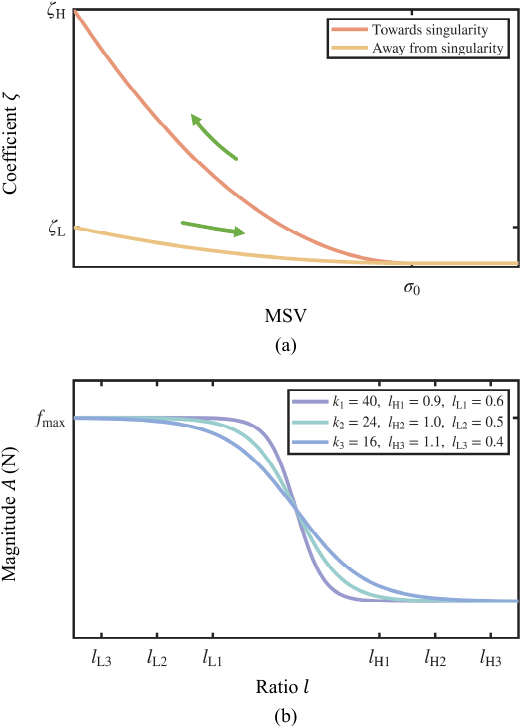}
	\caption{Influence of control parameters on dynamic force. (a) Admittance parameters. (b) Virtual force. The control parameters can be reasonably selected according to task requirements to adjust the magnitude and effective range of the dynamic force.}
	\label{fig3}
\end{figure}

Since the introduction of a virtual force injects energy into the human-robot system, the magnitude of the virtual force should be constrained within a reasonable range. A logistic function can be used to smoothly scale the magnitude of the virtual force \cite{ref26}, allowing the generated virtual force to effectively guide the user without replacing user control, which would reduce comfort and potentially introduce stability issues. The expression of the logistic function is given as:
\begin{equation}\label{eq11}	
	A=\frac{{{f}_{\max }}}{1+{{e}^{k(l-\frac{{{l}_{\mathrm{H}}}+{{l}_{\mathrm{L}}}}{2})}}}
\end{equation}
where ${f}_{\max}$ is the predefined maximum value of the virtual force. The parameter $k$ controls the sharpness of the function, determining the rate of change. The variable $l$ is defined as $l=\sigma_{\mathrm{m}}/\sigma_{0}$, where $\sigma_{\mathrm{m}}$ denotes the current MSV of the Jacobian matrix, and $\sigma_{0}$ is the set threshold. The constants ${l}_{\mathrm{H}}$ and ${l}_{\mathrm{L}}$ control the effective range of the virtual force in the robot workspace. The influence of control parameters on the dynamic force is illustrated in \hyperref[fig3]{Fig.~\ref*{fig3}}.

From \hyperref[fig3]{Fig.~\ref*{fig3}(b)}, it can be observed that when $l<{{l}_{\mathrm{L}}}$, the virtual force approaches the maximum value ${f}_{\max}$, and when $l>{{l}_{\mathrm{H}}}$, the virtual force tends towards zero. This indicates that the virtual force takes effect only within a predefined range. By adjusting the parameters ${l}_{\mathrm{H}}$ and ${l}_{\mathrm{L}}$, the effective range of the virtual force in the robot workspace can be controlled. The parameter $k$ determines the rate at which the virtual force changes and can be selected according to the requirements of the task.

To guide the user in moving the robot towards a configuration with optimized performance, the direction of the virtual force is aligned with the gradient direction of the performance index in Cartesian space. Since the performance index is a function of the joint positions of the robot and may not have an analytical form, a numerical strategy is adopted. The gradient of the performance index is approximated by computing the change in its value under small displacements in different directions, thereby reducing the computational cost. Specifically, the algorithm for one principal axis direction in Cartesian space is described, and computations in other directions can be carried out in the similar way. Let the current joint position of the robot be ${\mathbf{q}}_{0}$, the translational Jacobian matrix be ${\bar{\mathbf{J}}}_{\mathrm{t}}(\mathbf{q}_{0})$, and the performance index be $\sigma_{\mathrm{m}}$. Consider a small displacement $\delta \mathbf{X}={{\left[ \delta x,0,0 \right]}^{T}}$ along the $x$-direction in task space, and map this displacement into joint space using the translational Jacobian matrix:
\begin{equation}\label{eq12}	
	{{\hat{\mathbf{q}}}_{\pm }}={{\mathbf{q}}_{0}}\pm \bar{\mathbf{J}}_{\mathrm{t}}^{\dagger }({{\mathbf{q}}_{0}})\delta \mathbf{X}
\end{equation}

The translational Jacobian matrix $\bar{\mathbf{J}}({{\hat{\mathbf{q}}}_{\pm }})$ is computed based on ${{\hat{\mathbf{q}}}_{\pm }}$, and the corresponding MSV is denoted as ${{\hat{\sigma }}_{m\pm }}$. Based on the change in the performance index along the two positions, the gradient of the performance index in the $x$-direction is defined as:
\begin{equation}\label{eq13}	
	{{g}_{x}}=\left\{ \begin{array}{*{35}{l}}
		0, & (C1){{\gamma }_{+}}\le 0\wedge {{\gamma }_{-}}\le 0  \\
		{{\gamma }_{+}}, & (C2){{\gamma }_{+}}\ge {{\gamma }_{-}}\wedge \neg (C1)  \\
		-{{\gamma }_{-}}, & (C3)otherwise  \\
	\end{array} \right.
\end{equation}
where ${g}_{x}$ is the $x$-component of the performance index gradient $\mathbf{G}$. The difference between the calculated MSV and the current MSV is expressed as ${{\gamma }_{\pm }}={{\hat{\sigma }}_{m\pm }}-{{\sigma }_{\mathrm{m}}}$. When the performance indices at both displaced positions are smaller than the current value, it indicates that the performance index at the current position is at a local maximum along the $x$-direction, and no guidance force is needed in that direction. When at least one of the displacements leads to an increase in the performance index, the direction that results in the increase is selected as the gradient direction. Similarly, the $y$ and $z$ components of the gradient vector $\mathbf{G}$, denoted as ${g}_{y}$ and ${g}_{z}$, can be computed in the same way. The gradient vector $\mathbf{G}$ is then normalized as $\hat{\mathbf{G}}$ to obtain the unit guidance direction. Combining with \hyperref[eq11]{(\ref{eq11})}, the final generated virtual force is expressed as:
\begin{equation}\label{eq14}	
	{{\mathbf{F}}_{\mathrm{v}}}=A \hat{\mathbf{G}}
\end{equation}

\begin{figure}[pos=!t]
	\centering
	\includegraphics[width=\linewidth]{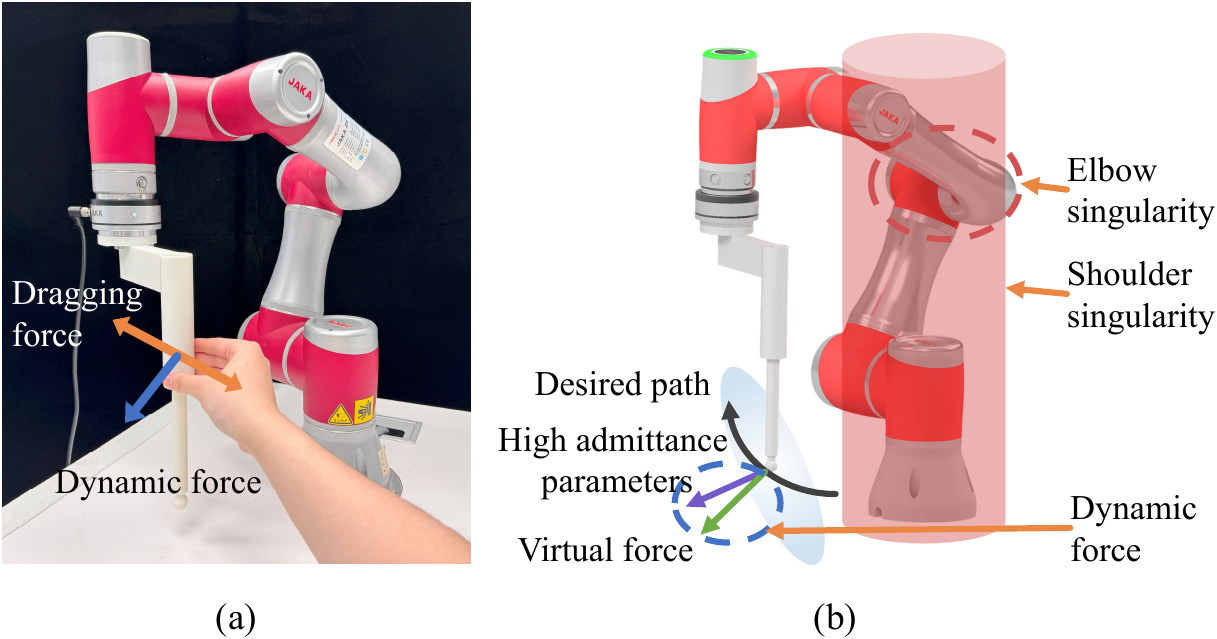}
	\caption{The mechanism of the dynamic force guidance. (a) Kinesthetic teaching with dynamic force guidance. (b) Control principle.}
	\label{fig4}
\end{figure}

The proposed method applies performance constraint based on variable admittance in the direction of the MSV and generates a virtual force in the direction of performance optimization, jointly guiding the user to improve the quality of the taught trajectory. The mechanism of the force guidance is illustrated in \hyperref[fig4]{Fig.~\ref*{fig4}}.

\subsection{Stability Analysis}

In physical human-robot interaction, the system is typically required to maintain energy passivity to ensure stability \cite{keemink2018admittance,chen2022human}. However, the proposed DFGT strategy introduces an active virtual force. This implies that energy is actively injected into the system, which inevitably breaks the passivity. Therefore, this paper conducts a stability analysis based on the bounded energy dissipation criterion. According to the proposed method, the admittance equation of the system in Cartesian space can be expressed as:
\begin{equation}\label{eq15}	
	\mathbf{M}_{\mathrm{d}}^{*}\overset{..}{\mathop{\mathbf{X}}}\,+\mathbf{C}_{\mathrm{d}}^{*}\overset{.}{\mathop{\mathbf{X}}}\,={{\mathbf{F}}_{\mathrm{h}}}+{{\mathbf{F}}_{\mathrm{v}}}
\end{equation}

The storage function of the system is defined as follows:
\begin{equation}\label{eq16}	
	V(t) = \frac{1}{2} \dot{\mathbf{X}}^T \mathbf{M}_0 \dot{\mathbf{X}}
\end{equation}
where ${{\mathbf{M}}_{0}}$ is the initial constant mass matrix. Taking the time derivative of the storage function yields:
\begin{equation}\label{eq17}	
	\dot{V}(t) = \dot{\mathbf{X}}^T \mathbf{M}_0 \ddot{\mathbf{X}}
\end{equation}

Substituting \hyperref[eq15]{(\ref{eq15})} into the above equation yields:
\begin{equation}\label{eq18}	
	\dot{V}(t)=-{{\overset{.}{\mathop{\mathbf{X}}}\,}^{T}}{{\mathbf{M}}_{0}}{{(\mathbf{M}_{\mathrm{d}}^{*})}^{-1}}\mathbf{C}_{\mathrm{d}}^{*}\overset{.}{\mathop{\mathbf{X}}}\,+{{\overset{.}{\mathop{\mathbf{X}}}\,}^{T}}{{\mathbf{M}}_{0}}{{(\mathbf{M}_{\mathrm{d}}^{*})}^{-1}}({{\mathbf{F}}_{\mathrm{h}}}+{{\mathbf{F}}_{\mathrm{v}}})
\end{equation}

Since the mass matrix and damping matrix are synchronously adjusted in the performance constraint module, these two time-varying matrices satisfy the following relationship:
\begin{equation}\label{eq19}	
	(\mathbf{M}_{\mathrm{d}}^*)^{-1} \mathbf{C}_{\mathrm{d}}^* = \mathbf{M}_0^{-1} \mathbf{C}_0
\end{equation}
where ${{\mathbf{C}}_{0}}$ is the initial constant damping matrix. Substituting \hyperref[eq19]{(\ref{eq19})} into \hyperref[eq18]{(\ref{eq18})}, the derivative of the storage function can be simplified as:
\begin{equation}\label{eq20}	
	\dot{V}(t)=-{{\overset{.}{\mathop{\mathbf{X}}}\,}^{T}}{{\mathbf{C}}_{0}}\overset{.}{\mathop{\mathbf{X}}}\,+{{\overset{.}{\mathop{\mathbf{X}}}\,}^{T}}\mathbf{W}({{\mathbf{F}}_{\mathrm{h}}}+{{\mathbf{F}}_{\mathrm{v}}})
\end{equation}
where $\mathbf{W} = \mathbf{M}_0 (\mathbf{M}_{\mathrm{d}}^*)^{-1}$ is the weight matrix. When the robot approaches a singularity, the eigenvalues of the matrix $\mathbf{W}$ fall within $(0, 1]$. This indicates that the time-varying mass matrix attenuates the injection of external energy in low-performance regions. When the user drags the robot near a singularity, the active virtual force is activated. Upper-bounding the derivative of the storage function yields:
\begin{equation}\label{eq21}	
	\dot{V}(t)\le -{{\lambda }_{\mathrm{min}}}({{\mathbf{C}}_{0}})\|\overset{.}{\mathop{\mathbf{X}}}\,{{\|}^{2}}+\|\overset{.}{\mathop{\mathbf{X}}}\,\|\|{{\mathbf{F}}_{\mathrm{h}}}\|+\|\overset{.}{\mathop{\mathbf{X}}}\,\|\|{{\mathbf{F}}_{\mathrm{v}}}\|
\end{equation}

To ensure energy dissipation during the teaching process, the total energy increase rate of the system should be less than the input power supplied by the user. Taking the difference between the bounded rate of energy change and the input power yields:
\begin{equation}\label{eq22}	
	\dot{V}(t)-\|\overset{.}{\mathop{\mathbf{X}}}\,\|\|{{\mathbf{F}}_{\mathrm{h}}}\|\le -{{\lambda }_{\mathrm{min}}}({{\mathbf{C}}_{0}})\|\overset{.}{\mathop{\mathbf{X}}}\,{{\|}^{2}}+\|\overset{.}{\mathop{\mathbf{X}}}\,\|\|{{\mathbf{F}}_{\mathrm{v}}}\|
\end{equation}

When the above expression is less than zero, the system exhibits energy dissipation behavior. From this, the convergence boundary condition of the system velocity can be derived as:
\begin{equation}\label{eq23}	
	\|\overset{.}{\mathop{\mathbf{X}}}\,\|>\frac{\|{{\mathbf{F}}_{\mathrm{v}}}\|}{{{\lambda }_{\mathrm{min}}}({{\mathbf{C}}_{0}})}
\end{equation}

Since the virtual force primarily serves an auxiliary guidance, its maximum magnitude ${f}_{\mathrm{max}}$ is selected conservatively. As the virtual force decays rapidly when the robot moves away from singularities, the corresponding velocity bound is significantly compressed. Although the introduction of the virtual force locally breaks the passivity of the system, once the system state exceeds this boundary, the energy stored in the system stops increasing and gradually decays.

When the user drags the robot away from the singularity, the virtual force term gradually decays to zero. The mass matrix simultaneously restores to its initial value. At this point, the derivative of the storage function degenerates to:
\begin{equation}\label{eq24}	
	\dot{V}(t)=-{{\overset{.}{\mathop{\mathbf{X}}}\,}^{T}}{{\mathbf{C}}_{0}}\overset{.}{\mathop{\mathbf{X}}}\,+{{\overset{.}{\mathop{\mathbf{X}}}\,}^{T}}{{\mathbf{F}}_{\mathrm{h}}}
\end{equation}

The system satisfies the following inequality with respect to the input power:
\begin{equation}\label{eq25}	
	\dot{V}(t)-{{\overset{.}{\mathop{\mathbf{X}}}\,}^{T}}{{\mathbf{F}}_{\mathrm{h}}}=-{{\overset{.}{\mathop{\mathbf{X}}}\,}^{T}}{{\mathbf{C}}_{0}}\overset{.}{\mathop{\mathbf{X}}}\,\le -{{\lambda }_{\mathrm{min}}}({{\mathbf{C}}_{0}})\|\overset{.}{\mathop{\mathbf{X}}}\,{{\|}^{2}}<0
\end{equation}

This indicates that as the robot moves away from the singularity, the system satisfies the passivity. The residual energy within the system will be completely dissipated once the user stops applying the dragging force. Therefore, the proposed DFGT strategy guarantees bounded energy dissipation near singularities and passivity in the normal interaction region, thereby ensuring overall stability.

\section{Experimental Setup}\label{sc3}
\begin{figure}[pos=!t]
	\centering
	\includegraphics[width=0.9\linewidth]{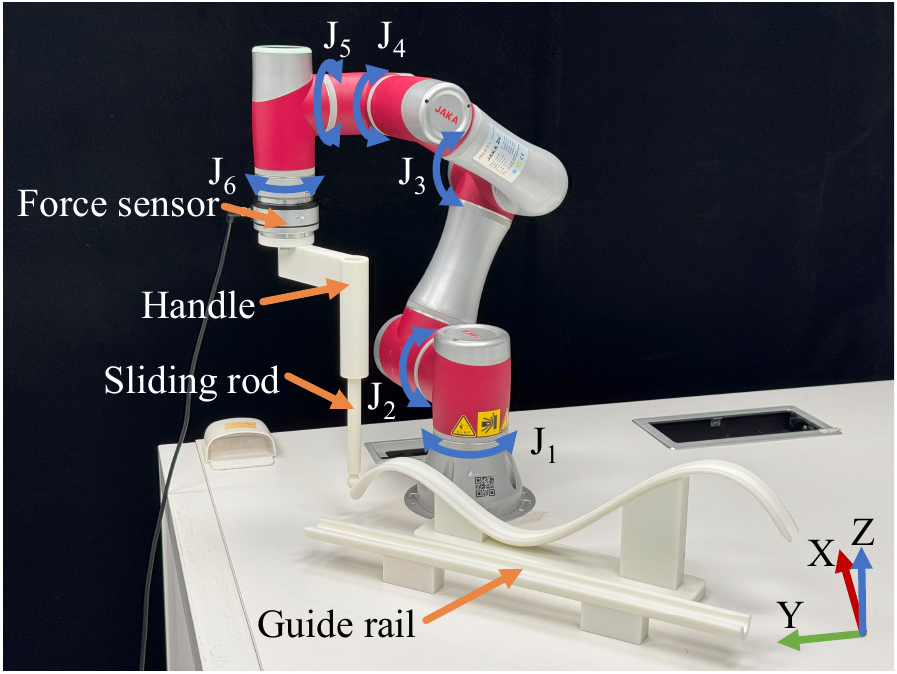}
	\caption{Experimental platform.}
	\label{fig5}
\end{figure}

The robot platform used in the experiment is a 6-DOF collaborative robot. A force sensor is installed at the end-effector to measure the operational force exerted by the user in real time. Below the sensor, a handle containing a sliding rod is mounted as the end-effector. To simulate the user dragging the robot along a desired path in Cartesian space, a guide rail is used in the experiment. The user interacts with the handle to control the motion of the robot and keeps the end of the rod within the guide rail to complete kinesthetic teaching along a specific path. The experimental platform is shown in \hyperref[fig5]{Fig.~\ref*{fig5}}.


\subsection{Preliminary}
The experiments use the MSV to evaluate the operational performance of the robot, and the threshold is set to 70\% of the maximum MSV within the robot workspace. The maximum value of the virtual force is set above the normal dragging force of the user during kinesthetic teaching to provide effective guidance while remaining comfortable and controllable. The control parameters used in the experiments are listed in \hyperref[tbl2]{Table~\ref*{tbl2}}.
\begin{table}[!t]
	\caption{Control parameters used in the experiment}\label{tbl2}
	\begin{center}
		\begin{tabularx}{0.9\linewidth}{XXXX}
			\toprule
			\multicolumn{2}{l}{Performance constraint} & \multicolumn{2}{l}{Performance optimization} \\
			\midrule
			Parameter & Value & Parameter & Value \\
			\midrule
			$\zeta_{\mathrm{H}}$ & 100 & $f_{\mathrm{max}}(\mathrm{N})$ & 15 \\
			$\zeta_{\mathrm{L}}$ & 15 & $k$ & 24 \\
			$\sigma_0$ & 96 & ${l}_{\mathrm{H}}$ & 1 \\
			\multicolumn{2}{l}{} & ${l}_{\mathrm{L}}$ & 0.5 \\
			\midrule
			\multicolumn{4}{l}{Admittance parameters} \\
			\midrule
			Parameter & \multicolumn{3}{X}{Value} \\
			$\mathbf{M}_{\mathrm{d}}$ & \multicolumn{3}{X}{$diag(16,16,16)\ [\mathrm{kg}]$} \\
			$\mathbf{C}_{\mathrm{d}}$ & \multicolumn{3}{X}{$diag(120,120,120)\ [\mathrm{Ns/m}]$} \\
			\bottomrule
		\end{tabularx}
	\end{center}
\end{table}

During kinesthetic teaching, the user may drag the robot into low-performance regions near singularities, resulting in degraded operational performance. The operational performance of the robot is analyzed using the Jacobian matrix. By setting $\det \mathbf{J}=0$, three types of singularities for the selected collaborative robot can be identified: (a) elbow singularity, (b) wrist singularity, and (c) shoulder singularity, where the origin of the fifth axis of the robot lies on a cylindrical surface centered at the base coordinate origin, as shown in \hyperref[fig4]{Fig.~\ref*{fig4}(b)}. Throughout the entire experiment, the end-effector maintains a posture perpendicular to the horizontal plane. Therefore, only elbow singularities and shoulder singularities are involved.

\subsection{Experimental Tasks}
The proposed method is implemented in the controller with a control frequency of 125 Hz. On the basis of the admittance control framework, the performance constraint modifies the admittance parameters to suppress motion towards singularities, and the performance optimization introduces an additional virtual force to guide the robot away from singularities. To verify its effectiveness, three comparative control strategies are designed. The first strategy is Direct Teaching (DT), which serves as a baseline for comparison and does not incorporate any additional mechanisms. When the robot reaches a singularity, joint velocities change drastically. In such cases, the robot automatically switches to a protection mode and terminates the teaching process. The second strategy is Variable Admittance Teaching (VAT), which applies only the performance constraint. When the robot is detected to be in a low-performance region, the controller dynamically increases the admittance parameters in the direction of the MSV to suppress further motion. The third strategy is Dynamic Force Guided Teaching (DFGT), which integrates performance constraint and optimization mechanisms by simultaneously adjusting the admittance parameters and introducing the virtual force to assist the user in naturally avoiding singularities.

\begin{figure*}[pos=!t]
	\centering
	\includegraphics[width=0.95\linewidth]{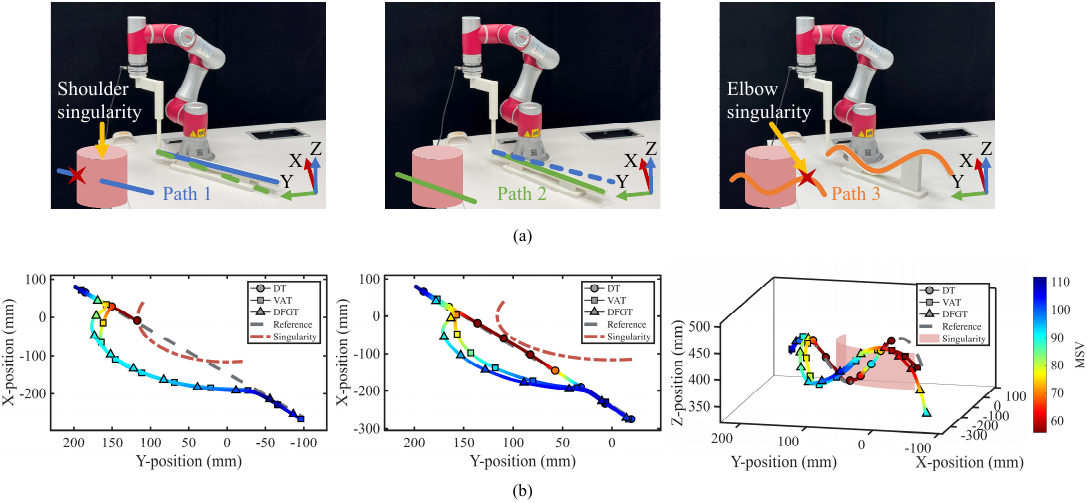}
	\caption{Experimental setup and performance indices. (a) Guide rails for three typical paths. (b) Taught trajectories and corresponding performance indices.}
	\label{fig6}
\end{figure*}

Under the three control strategies, the user manually guides the robot end-effector along the rail, gradually approaching the low-performance regions near singularities. To comprehensively evaluate the effectiveness of the three strategies, three typical paths are designed for the experiments: a straight path that reaches the shoulder singularity, a straight path that approaches but does not reach the shoulder singularity, and a curved path that reaches the elbow singularity. The experimental setup is shown in \hyperref[fig6]{Fig.~\ref*{fig6}(a)}. As the end-effector gradually approaches the singularity, the variable admittance and virtual force progressively take effect, providing effective suppression and guidance during kinesthetic teaching.

In addition, to evaluate the impact of different control strategies on the quality of the taught trajectories, this study recruits ten healthy participants for the experiment (age: 23–44 years, height: 164–185 cm, weight: 50–90 kg). Considering that passing through a singularity leads to an operation interruption, the experiment selects Path 2 for analysis. Each subject repeats the teaching task three times under each of the three control strategies, collecting a total of 90 valid taught trajectories. The length and the MSV of each trajectory are then obtained. To investigate the relationship between the operational performance of the robot and the work efficiency, all trajectories are played back. The maximum rated velocities of the robot joints are $\dot{\mathbf{q}}_{\mathrm{lim}} = [180, 180, 180, 220, 220, 220] (^{\circ}/s)$. To ensure system safety, the velocity limits of all joints are constrained to 10\% of their maximum rated velocities. The actual execution time $T_{\mathrm{ex}}$ of each trajectory is recorded.

\section{Results}\label{sc4}
To verify the effectiveness of the proposed method in kinesthetic teaching tasks, comparison experiments are conducted based on the three typical paths. During the experiments, key data including the end-effector position, MSV, joint velocities, and virtual force are recorded for comparative analysis of different control strategies. To evaluate the quality of the taught trajectories, multiple groups of experiments are conducted, and the statistical results are compared using t-tests.

\subsection{Validation of Dynamic Force Guidance}
\begin{figure*}[pos=!t]
	\centering
	\includegraphics[width=\linewidth]{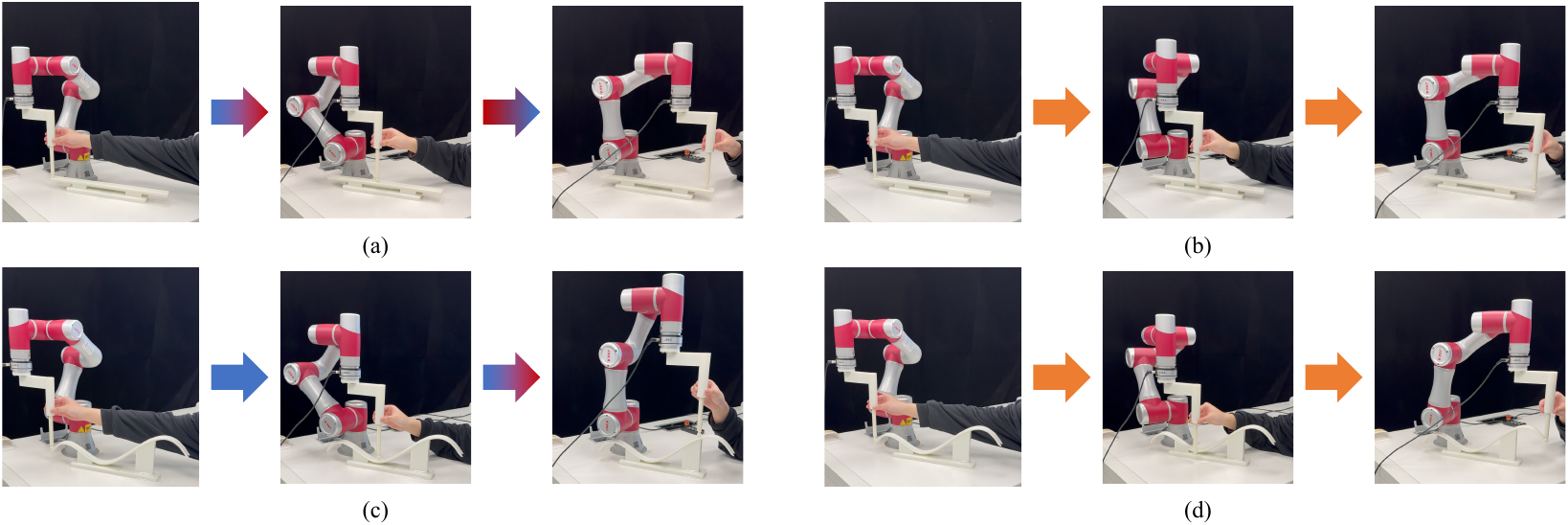}
	\caption{Real-world kinesthetic teaching demonstrations. (a) DT strategy on Path 2. (b) DFGT strategy on Path 2. (c) DT strategy on Path 3. (d) DFGT strategy on Path 3. The situation for Path 1 is similar to that of Path 2. The DT strategy reaches the singularity under both Path 1 and Path 3, leading to the termination of the kinesthetic teaching process.}
	\label{fig7}
\end{figure*}

The spatial trajectories recorded under different control strategies, along with the corresponding performance indices, are shown in \hyperref[fig6]{Fig.~\ref*{fig6}(b)}. The coordinate system in the figure is aligned with the base coordinate frame of the robot. It can be observed that in the DT strategy, the robot moves along the rail, and the performance index drops significantly when approaching the singularity. Upon reaching the singularity, the robot automatically enters protection mode and terminates the kinesthetic teaching process. In the VAT strategy, motion towards the singularity is noticeably suppressed. As a result, the user tends to guide the robot around the low-performance regions, keeping the operational performance of the robot at a relatively high level. Building on this, the introduction of virtual force further enhances the operational performance of the robot while guiding the user in the direction of performance optimization. The real-world experimental demonstrations are shown in \hyperref[fig7]{Fig.~\ref*{fig7}}.

\begin{figure*}[pos=!t]
	\centering
	\includegraphics[width=\linewidth]{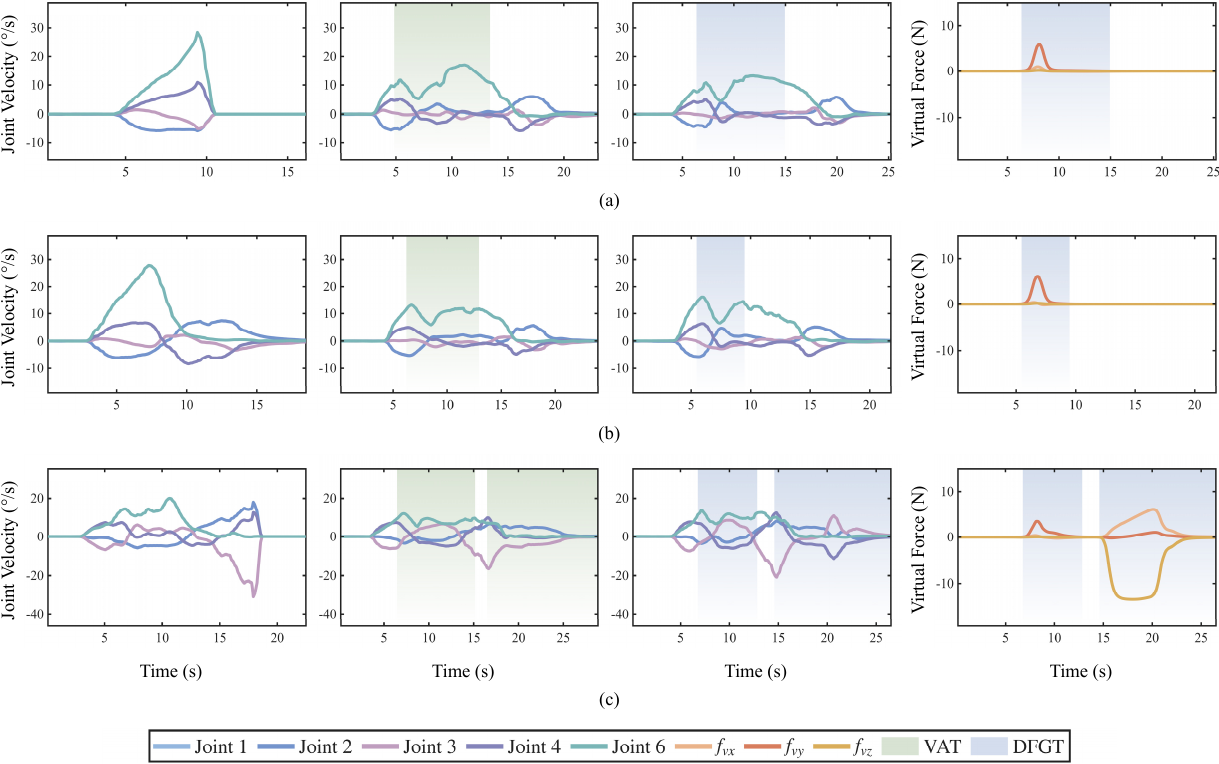}
	\caption{Joint velocity and virtual force under different control strategies. (a) Direct Teaching. (b) Variable Admittance Teaching. (c) Dynamic Force Guided Teaching. Each row corresponds to one predefined path.}
	\label{fig8}
\end{figure*}

\hyperref[fig8]{Fig.~\ref*{fig8}} further presents the joint velocity responses under the three strategies and the generated virtual force. It can be observed that when the robot is moved towards a singularity in the DT strategy, joint velocities increase sharply. This may lead to system instability and impact loads, posing safety risks. In contrast, the DFGT strategy effectively suppresses the divergence of joint velocities caused by Jacobian matrix degradation and enhances the operational performance of the robot. From the virtual force results, it can be observed that as the robot gradually approaches the low-performance regions near singularities, the virtual force emerges progressively and guides the user to optimize the operational performance of the robot. Specifically, at the end of Path 3, although damping significantly suppresses motion towards the singularity, its passive nature causes the robot to remain in a region with relatively lower operational performance once the user stops applying the dragging force. When virtual force is introduced, it further guides the robot to move in the direction of increasing performance index, thereby optimizing the operational performance of the robot. The direction of the virtual force aligns with the gradient of the MSV, and its magnitude can be dynamically adjusted. By limiting the virtual force within a reasonable range, the system provides effective active guidance without overriding the control of the user. As a result, the generated virtual force exhibits strong guidance capability.

\subsection{Trajectory Quality Evaluation}
\begin{figure*}[pos=!t]
	\centering
	\includegraphics[width=0.9\linewidth]{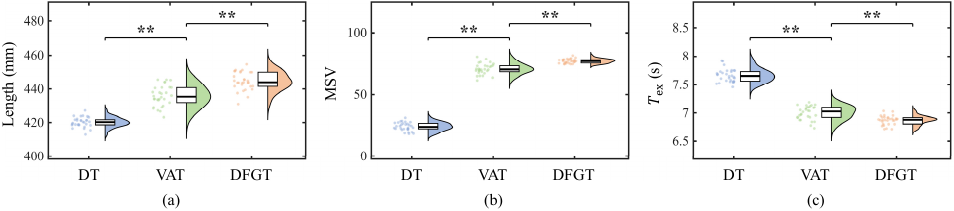}
	\caption{Statistical results of evaluation metrics for different control strategies on Path 2. (a) Trajectory length. (b) Minimum Singular Value. (c) Actual execution time. Asterisks indicate statistical significance: $^{**} p < 0.001$.}
	\label{fig9}
\end{figure*}

To evaluate the quality of the taught trajectories under different control strategies, 90 valid taught trajectories are played back. \hyperref[fig9]{Fig.~\ref*{fig9}} presents the statistical results of the evaluation metrics for different control strategies, with their mean values listed in \hyperref[tbl3]{Table~\ref*{tbl3}}. The results show that although the DT strategy produces the shortest taught trajectory, its $T_{\mathrm{ex}}$ is the longest among the three strategies, indicating lower work efficiency. In contrast, the introduction of variable admittance and virtual force leads to slightly longer taught trajectories, but the improvement in the operational performance of the robot results in more well-conditioned joint motion, which in turn reduces $T_{\mathrm{ex}}$. Compared to the DT strategy, the $T_{\mathrm{ex}}$ under the DFGT strategy decreases by 10.30\%.
\begin{table}[!t]
	\caption{Evaluation Metrics under Different Control Strategies on Path 2}
	\label{tbl3}
	\begin{center}
		\begin{tabularx}{0.95\linewidth}{Xp{1.5cm}XXX}
			\toprule
			Strategy & Length (mm) & MSV & $T_{\mathrm{ex}}$ (s) &  $\overline{F}_{\mathrm{h}}^{\max}$ (N)\\
			\midrule
			DT & 420.22 & 23.92 & 7.66 &  4.76\\			
			VAT & 435.80 & 70.83 & 7.00 &  8.83\\
			DFGT & 444.62 & 77.15 & 6.87 &  6.96\\
			\bottomrule
		\end{tabularx}
	\end{center}
\end{table}

\begin{figure}[pos=!t]
	\centering
	\includegraphics[width=0.8\linewidth]{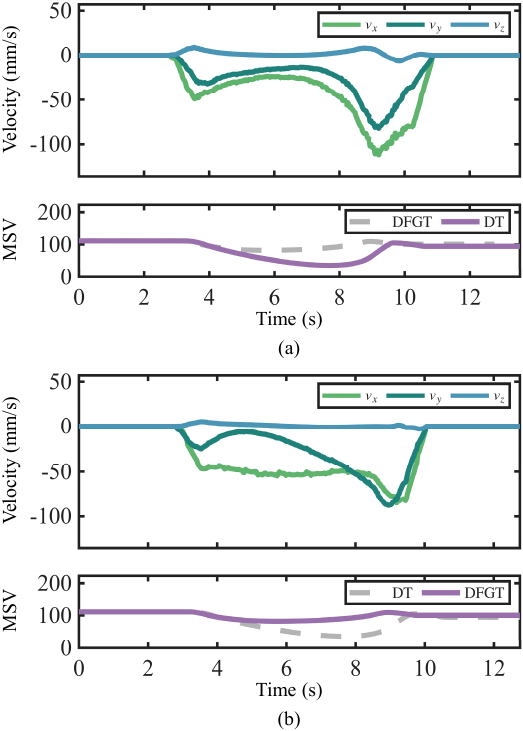}
	\caption{End-effector velocity and MSV over time during the playback process. (a) Direct Teaching. (b) Dynamic Force Guided Teaching.}
	\label{fig10}
\end{figure}

\hyperref[fig10]{Fig.~\ref*{fig10}} reveals the reason for the differences in actual execution time. It illustrates the relationship between the end-effector velocity and the MSV on Path 2 under the DT and DFGT strategies. When the robot approaches the singularity, the MSV decreases. Once the joint velocities reach their saturation limits, the end-effector velocity mapped through the Jacobian matrix decreases significantly. When the robot moves away from the singularity, the MSV gradually increases. The operational performance of the robot gradually recovers, and the end-effector velocity increases accordingly. By actively guiding the user away from low-performance regions near singularities, the DFGT strategy optimizes the actual execution time. This result indicates that the DFGT strategy can effectively enhance work efficiency.

\subsection{Discussion}
The proposed method integrates performance constraint based on variable admittance with performance optimization based on virtual force, effectively improving the operational performance of the robot during kinesthetic teaching. Experimental results show that the performance constraint can maintain the operational performance of the robot above a predefined threshold and prevent it from entering low-performance regions near singularities. On this basis, the introduction of the virtual force actively guides the user to move the robot towards directions that optimize the performance. As shown by the average maximum dragging force $\overline{F}_{\mathrm{h}}^{\max}$ in \hyperref[tbl3]{Table~\ref*{tbl3}}., users tend to apply a larger force under the VAT strategy due to the damping effect when it becomes difficult to drag the robot towards singularities. In contrast, the virtual force in the DFGT strategy effectively reduces user effort and makes the teaching process more comfortable. Due to the effect of the performance constraint, the magnitude of the virtual force can be reasonably limited within a bounded range. This is because the virtual force does not impose a strict constraint on the motion of the robot, but serves as auxiliary feedback for performance optimization. Compared with unbounded virtual force methods, this approach effectively avoids system instability caused by excessive force. Compared to the DLS method, the proposed method does not sacrifice the accuracy of the inverse kinematic solution and therefore maintains better consistency in motion. In addition, this paper mainly considers the case where the end-effector maintains a constant orientation. However, the proposed method is extensible and can be applied to more complex tasks involving both translation and rotation, given that appropriate performance indices are defined.

It is worth noting that the introduction of dynamic force guidance causes the actual taught trajectory to deviate from the desired path, resulting in an increase in trajectory length. Trajectory quality is analyzed based on $T_{\mathrm{ex}}$. Results show that although the trajectory length increases under the effect of dynamic guidance force, the $T_{\mathrm{ex}}$ required to execute the trajectory is reduced due to improved operational performance. This indicates that optimizing operational performance contributes to improving the work efficiency.

\section{Conclusion}\label{sc5}
This paper proposes a control framework for kinesthetic teaching of collaborative robots. By combining performance constraint based on variable admittance with performance optimization through virtual force, an online dynamic force guidance mechanism is constructed. The method increases admittance parameters to suppress motion towards low-performance directions during kinesthetic teaching, while virtual force guides the user to optimize the operational performance of the robot. Experimental results show that the proposed method demonstrates effective guidance across various paths. In addition, trajectory playback experiments are conducted. Statistical results reveal the benefits of enhancing the operational performance of the robot on improving the work efficiency. This provides a theoretical basis for reducing production takt time in industrial deployment.

Meanwhile, there are still several aspects that deserve further research. Exploring new performance indices is essential to extend the method to more complex tasks involving both translation and rotation. In addition, future work will integrate subjective evaluation metrics to quantitatively analyze the differences between the proposed method and other existing methods.

\printcredits

\bibliographystyle{cas-model2-names}

\bibliography{references}

\begin{thebibliography}{34}
\expandafter\ifx\csname natexlab\endcsname\relax\def\natexlab#1{#1}\fi
\providecommand{\url}[1]{\texttt{#1}}
\providecommand{\href}[2]{#2}
\providecommand{\path}[1]{#1}
\providecommand{\DOIprefix}{}
\providecommand{\ArXivprefix}{arXiv:}
\providecommand{\URLprefix}{}
\providecommand{\Pubmedprefix}{pmid:}
\providecommand{\doi}[1]{\href{https://doi.org/#1}{https://doi.org/#1}}
\providecommand{\Pubmed}[1]{\href{pmid:#1}{\path{#1}}}
\providecommand{\bibinfo}[2]{#2}
\ifx\xfnm\relax \def\xfnm[#1]{\unskip,\space#1}\fi
\bibitem[{Zaatari et~al.(2019)Zaatari, Marei, Li and Usman}]{ref1}
\bibinfo{author}{Zaatari SE}, \bibinfo{author}{Marei M}, \bibinfo{author}{Li
  W}, \bibinfo{author}{Usman Z}. \bibinfo{title}{Cobot programming for
  collaborative industrial tasks: An overview}.
\newblock \bibinfo{journal}{Robot Auton Syst}
  \bibinfo{year}{2019}\bibinfo{volume}{;116}:\bibinfo{pages}{162--180}.
\newblock \DOIprefix\doi{10.1016/j.robot.2019.03.003}.
\bibitem[{Deniša et~al.(2016)Deniša, Gams, Ude and Petrič}]{ref2}
\bibinfo{author}{Deniša M}, \bibinfo{author}{Gams A}, \bibinfo{author}{Ude A},
  \bibinfo{author}{Petrič T}. \bibinfo{title}{Learning compliant movement
  primitives through demonstration and statistical generalization}.
\newblock \bibinfo{journal}{IEEE-ASME Trans Mechatron}
  \bibinfo{year}{2016};21\bibinfo{volume}{(5)}:\bibinfo{pages}{2581--2594}.
\newblock \DOIprefix\doi{10.1109/TMECH.2015.2510165}.
\bibitem[{Kronander et~al.(2015)Kronander, Khansari and Billard}]{ref3}
\bibinfo{author}{Kronander K}, \bibinfo{author}{Khansari M},
  \bibinfo{author}{Billard A}. \bibinfo{title}{Incremental motion learning with
  locally modulated dynamical systems}.
\newblock \bibinfo{journal}{Robot Auton Syst}
  \bibinfo{year}{2015}\bibinfo{volume}{;70}:\bibinfo{pages}{52--62}.
\newblock \DOIprefix\doi{10.1016/j.robot.2015.03.010}.
\bibitem[{Vogt et~al.(2017)Vogt, Stepputtis, Grehl, Jung and Ben~Amor}]{ref4}
\bibinfo{author}{Vogt D}, \bibinfo{author}{Stepputtis S},
  \bibinfo{author}{Grehl S}, \bibinfo{author}{Jung B},
  \bibinfo{author}{Ben~Amor H}. \bibinfo{title}{A system for learning
  continuous human-robot interactions from human-human demonstrations}.  In:
  \bibinfo{booktitle}{Proceedings of the IEEE International Conference on
  Robotics and Automation}. \bibinfo{organization}{IEEE}; \bibinfo{year}{2017},
  p. \bibinfo{pages}{2882--2889}.
\newblock \DOIprefix\doi{10.1109/ICRA.2017.7989334}.
\bibitem[{Meattini et~al.(2025)Meattini, Govoni, Galassi, Chiaravalli, Palli
  and Melchiorri}]{ref31}
\bibinfo{author}{Meattini R}, \bibinfo{author}{Govoni A},
  \bibinfo{author}{Galassi K}, \bibinfo{author}{Chiaravalli D},
  \bibinfo{author}{Palli G}, \bibinfo{author}{Melchiorri C}.
  \bibinfo{title}{Programming robot interaction behavior during kinesthetic
  teaching exploiting semg-based interfacing and vibrotactile feedback}.
\newblock \bibinfo{journal}{IEEE-ASME Trans Mechatron}
  \bibinfo{year}{2025};30\bibinfo{volume}{(5)}:\bibinfo{pages}{4011--4022}.
\newblock \DOIprefix\doi{10.1109/TMECH.2025.3603402}.
\bibitem[{Sasabuchi et~al.(2021)Sasabuchi, Wake and Ikeuchi}]{ref6}
\bibinfo{author}{Sasabuchi K}, \bibinfo{author}{Wake N},
  \bibinfo{author}{Ikeuchi K}. \bibinfo{title}{Task-oriented motion mapping on
  robots of various configuration using body role division}.
\newblock \bibinfo{journal}{IEEE Robot Autom Lett}
  \bibinfo{year}{2021};6\bibinfo{volume}{(2)}:\bibinfo{pages}{413--420}.
\newblock \DOIprefix\doi{10.1109/LRA.2020.3044029}.
\bibitem[{Kang et~al.(2019)Kang, Oh, Seo, Kim and Choi}]{ref30}
\bibinfo{author}{Kang G}, \bibinfo{author}{Oh HS}, \bibinfo{author}{Seo JK},
  \bibinfo{author}{Kim U}, \bibinfo{author}{Choi HR}. \bibinfo{title}{Variable
  admittance control of robot manipulators based on human intention}.
\newblock \bibinfo{journal}{IEEE-ASME Trans Mechatron}
  \bibinfo{year}{2019};24\bibinfo{volume}{(3)}:\bibinfo{pages}{1023--1032}.
\newblock \DOIprefix\doi{10.1109/TMECH.2019.2910237}.
\bibitem[{Lafleche et~al.(2019)Lafleche, Saunderson and Nejat}]{ref7}
\bibinfo{author}{Lafleche JF}, \bibinfo{author}{Saunderson S},
  \bibinfo{author}{Nejat G}. \bibinfo{title}{Robot cooperative behavior
  learning using single-shot learning from demonstration and parallel hidden
  markov models}.
\newblock \bibinfo{journal}{IEEE Robot Autom Lett}
  \bibinfo{year}{2019};4\bibinfo{volume}{(2)}:\bibinfo{pages}{193--200}.
\newblock \DOIprefix\doi{10.1109/LRA.2018.2885584}.
\bibitem[{Hogan(1984)}]{ref8}
\bibinfo{author}{Hogan N}. \bibinfo{title}{Impedance control: An approach to
  manipulation}.  In: \bibinfo{booktitle}{1984 American Control Conference}.
  \bibinfo{organization}{IEEE}; \bibinfo{year}{1984}, p.
  \bibinfo{pages}{304--313}.
\newblock \DOIprefix\doi{10.23919/ACC.1984.4788393}.
\bibitem[{Han et~al.(2024)Han, Zhao, Huang and Xu}]{han2024variable}
\bibinfo{author}{Han L}, \bibinfo{author}{Zhao L}, \bibinfo{author}{Huang Y},
  \bibinfo{author}{Xu W}. \bibinfo{title}{Variable admittance control for safe
  physical human–robot interaction considering intuitive human intention}.
\newblock \bibinfo{journal}{Mechatronics}
  \bibinfo{year}{2024}\bibinfo{volume}{;97}:\bibinfo{pages}{103098}.
\newblock \DOIprefix\doi{10.1016/j.mechatronics.2023.103098}.
\bibitem[{Ravichandar et~al.(2020)Ravichandar, Polydoros, Chernova and
  Billard}]{ref9}
\bibinfo{author}{Ravichandar H}, \bibinfo{author}{Polydoros AS},
  \bibinfo{author}{Chernova S}, \bibinfo{author}{Billard A}.
  \bibinfo{title}{Recent advances in robot learning from demonstration}.
\newblock \bibinfo{journal}{Annu Rev Contr Robot Autonom Syst}
  \bibinfo{year}{2020};3\bibinfo{volume}{(1)}:\bibinfo{pages}{297--330}.
\newblock \DOIprefix\doi{10.1146/annurev-control-100819-063206}.
\bibitem[{Nguyen et~al.(2023)Nguyen, Campeau-Lecours and Gosselin}]{ref32}
\bibinfo{author}{Nguyen TS}, \bibinfo{author}{Campeau-Lecours A},
  \bibinfo{author}{Gosselin C}. \bibinfo{title}{Physical human–robot
  interaction using a macro–mini robotic system}.
\newblock \bibinfo{journal}{IEEE-ASME Trans Mechatron}
  \bibinfo{year}{2023};28\bibinfo{volume}{(6)}:\bibinfo{pages}{3398--3409}.
\newblock \DOIprefix\doi{10.1109/TMECH.2023.3267781}.
\bibitem[{Lee et~al.(2020)Lee, Lee, Park and Chung}]{ref10}
\bibinfo{author}{Lee D}, \bibinfo{author}{Lee W}, \bibinfo{author}{Park J},
  \bibinfo{author}{Chung WK}. \bibinfo{title}{Task space control of articulated
  robot near kinematic singularity: Forward dynamics approach}.
\newblock \bibinfo{journal}{IEEE Robot Autom Lett}
  \bibinfo{year}{2020};5\bibinfo{volume}{(2)}:\bibinfo{pages}{752--759}.
\newblock \DOIprefix\doi{10.1109/LRA.2020.2965071}.
\bibitem[{Pulloquinga et~al.(2023)Pulloquinga, Escarabajal, Vallés,
  Díaz-Rodríguez, Mata and Ángel Valera}]{pulloquinga2023admittance}
\bibinfo{author}{Pulloquinga JL}, \bibinfo{author}{Escarabajal RJ},
  \bibinfo{author}{Vallés M}, \bibinfo{author}{Díaz-Rodríguez M},
  \bibinfo{author}{Mata V}, \bibinfo{author}{Ángel Valera}.
  \bibinfo{title}{Admittance controller complemented with real-time singularity
  avoidance for rehabilitation parallel robots}.
\newblock \bibinfo{journal}{Mechatronics}
  \bibinfo{year}{2023}\bibinfo{volume}{;94}:\bibinfo{pages}{103017}.
\newblock \DOIprefix\doi{10.1016/j.mechatronics.2023.103017}.
\bibitem[{Wampler(1986)}]{ref11}
\bibinfo{author}{Wampler CW}. \bibinfo{title}{Manipulator inverse kinematic
  solutions based on vector formulations and damped least-squares methods}.
\newblock \bibinfo{journal}{IEEE Trans Syst Man Cybern}
  \bibinfo{year}{1986};16\bibinfo{volume}{(1)}:\bibinfo{pages}{93--101}.
\newblock \DOIprefix\doi{10.1109/TSMC.1986.289285}.
\bibitem[{Caccavale et~al.(1997)Caccavale, Chiaverini and Siciliano}]{ref33}
\bibinfo{author}{Caccavale F}, \bibinfo{author}{Chiaverini S},
  \bibinfo{author}{Siciliano B}. \bibinfo{title}{Second-order kinematic control
  of robot manipulators with jacobian damped least-squares inverse: theory and
  experiments}.
\newblock \bibinfo{journal}{IEEE-ASME Trans Mechatron}
  \bibinfo{year}{1997};2\bibinfo{volume}{(3)}:\bibinfo{pages}{188--194}.
\newblock \DOIprefix\doi{10.1109/3516.622971}.
\bibitem[{Carmichael et~al.(2017)Carmichael, Liu and Waldron}]{ref12}
\bibinfo{author}{Carmichael MG}, \bibinfo{author}{Liu D},
  \bibinfo{author}{Waldron KJ}. \bibinfo{title}{A framework for
  singularity-robust manipulator control during physical human-robot
  interaction}.
\newblock \bibinfo{journal}{Int J Robot Res}
  \bibinfo{year}{2017};36\bibinfo{volume}{(5-7)}:\bibinfo{pages}{861--876}.
\newblock \DOIprefix\doi{10.1177/0278364917698748}.
\bibitem[{Dimeas et~al.(2016)Dimeas, Moulianitis, Papakonstantinou and
  Aspragathos}]{ref13}
\bibinfo{author}{Dimeas F}, \bibinfo{author}{Moulianitis VC},
  \bibinfo{author}{Papakonstantinou C}, \bibinfo{author}{Aspragathos N}.
  \bibinfo{title}{Manipulator performance constraints in cartesian admittance
  control for human-robot cooperation}.  In: \bibinfo{booktitle}{Proceedings of
  the IEEE International Conference on Robotics and Automation}.
  \bibinfo{organization}{IEEE}; \bibinfo{year}{2016}, p.
  \bibinfo{pages}{3049--3054}.
\newblock \DOIprefix\doi{10.1109/ICRA.2016.7487469}.
\bibitem[{Dimeas et~al.(2018)Dimeas, Moulianitis and Aspragathos}]{ref14}
\bibinfo{author}{Dimeas F}, \bibinfo{author}{Moulianitis VC},
  \bibinfo{author}{Aspragathos N}. \bibinfo{title}{Manipulator performance
  constraints in human-robot cooperation}.
\newblock \bibinfo{journal}{Robot Comput-Integr Manuf}
  \bibinfo{year}{2018}\bibinfo{volume}{;50}:\bibinfo{pages}{222--233}.
\newblock \DOIprefix\doi{10.1016/j.rcim.2017.09.015}.
\bibitem[{Kim et~al.(2021)Kim, Jie, Kim and Lee}]{ref34}
\bibinfo{author}{Kim J}, \bibinfo{author}{Jie W}, \bibinfo{author}{Kim H},
  \bibinfo{author}{Lee MC}. \bibinfo{title}{Modified configuration control with
  potential field for inverse kinematic solution of redundant manipulator}.
\newblock \bibinfo{journal}{IEEE-ASME Trans Mechatron}
  \bibinfo{year}{2021};26\bibinfo{volume}{(4)}:\bibinfo{pages}{1782--1790}.
\newblock \DOIprefix\doi{10.1109/TMECH.2021.3077914}.
\bibitem[{Yoshikawa(1985)}]{ref15}
\bibinfo{author}{Yoshikawa T}. \bibinfo{title}{Manipulability of robotic
  mechanisms}.
\newblock \bibinfo{journal}{Int J Robot Res}
  \bibinfo{year}{1985};4\bibinfo{volume}{(2)}:\bibinfo{pages}{3--9}.
\newblock \DOIprefix\doi{10.1177/027836498500400201}.
\bibitem[{Salisbury and Craig(1982)}]{ref16}
\bibinfo{author}{Salisbury JK}, \bibinfo{author}{Craig JJ}.
  \bibinfo{title}{Articulated hands: Force control and kinematic issues}.
\newblock \bibinfo{journal}{Int J Robot Res}
  \bibinfo{year}{1982};1\bibinfo{volume}{(1)}:\bibinfo{pages}{4--17}.
\newblock \DOIprefix\doi{10.1177/027836498200100102}.
\bibitem[{Klein and Blaho(1987)}]{ref17}
\bibinfo{author}{Klein CA}, \bibinfo{author}{Blaho BE}.
  \bibinfo{title}{Dexterity measures for the design and control of
  kinematically redundant manipulators}.
\newblock \bibinfo{journal}{Int J Robot Res}
  \bibinfo{year}{1987};6\bibinfo{volume}{(2)}:\bibinfo{pages}{72--83}.
\newblock \DOIprefix\doi{10.1177/027836498700600206}.
\bibitem[{Patel and Sobh(2015)}]{ref18}
\bibinfo{author}{Patel S}, \bibinfo{author}{Sobh T}.
  \bibinfo{title}{Manipulator performance measures-a comprehensive literature
  survey}.
\newblock \bibinfo{journal}{J Intell Robot Syst}
  \bibinfo{year}{2015}\bibinfo{volume}{;77}:\bibinfo{pages}{547--570}.
\newblock \DOIprefix\doi{10.1007/s10846-014-0024-y}.
\bibitem[{Schwartz et~al.(2002)Schwartz, Manseur and Doty}]{ref19}
\bibinfo{author}{Schwartz E}, \bibinfo{author}{Manseur R},
  \bibinfo{author}{Doty K}. \bibinfo{title}{Noncommensurate systems in
  robotics}.
\newblock \bibinfo{journal}{Int J Robot Autom}
  \bibinfo{year}{2002};17\bibinfo{volume}{(2)}:\bibinfo{pages}{86--92}.
\bibitem[{Yoshikawa(1990)}]{ref20}
\bibinfo{author}{Yoshikawa T}. \bibinfo{title}{Translational and rotational
  manipulability of robotic manipulators}.  In: \bibinfo{booktitle}{Proceedings
  of the American Control Conference}. \bibinfo{organization}{IEEE};
  \bibinfo{year}{1990}, p. \bibinfo{pages}{228--233}.
\newblock \DOIprefix\doi{10.23919/ACC.1990.4790733}.
\bibitem[{Cardou et~al.(2010)Cardou, Bouchard and Gosselin}]{ref21}
\bibinfo{author}{Cardou P}, \bibinfo{author}{Bouchard S},
  \bibinfo{author}{Gosselin C}. \bibinfo{title}{Kinematic-sensitivity indices
  for dimensionally nonhomogeneous jacobian matrices}.
\newblock \bibinfo{journal}{IEEE Trans Robot}
  \bibinfo{year}{2010};26\bibinfo{volume}{(1)}:\bibinfo{pages}{166--173}.
\newblock \DOIprefix\doi{10.1109/TRO.2009.2037252}.
\bibitem[{Mansouri and Ouali(2011)}]{ref22}
\bibinfo{author}{Mansouri I}, \bibinfo{author}{Ouali M}. \bibinfo{title}{The
  power manipulability – a new homogeneous performance index of robot
  manipulators}.
\newblock \bibinfo{journal}{Robot Comput-Integr Manuf}
  \bibinfo{year}{2011};27\bibinfo{volume}{(2)}:\bibinfo{pages}{434--449}.
\newblock \DOIprefix\doi{10.1016/j.rcim.2010.09.004}.
\bibitem[{Yoshikawa(1985)}]{ref24}
\bibinfo{author}{Yoshikawa T}. \bibinfo{title}{Dynamic manipulability of robot
  manipulators}.
\newblock \bibinfo{journal}{Trans Soc Instrum Control Eng}
  \bibinfo{year}{1985};21\bibinfo{volume}{(9)}:\bibinfo{pages}{970--975}.
\newblock \DOIprefix\doi{10.9746/sicetr1965.21.970}.
\bibitem[{Ficuciello et~al.(2015)Ficuciello, Villani and Siciliano}]{ref25}
\bibinfo{author}{Ficuciello F}, \bibinfo{author}{Villani L},
  \bibinfo{author}{Siciliano B}. \bibinfo{title}{Variable impedance control of
  redundant manipulators for intuitive human–robot physical interaction}.
\newblock \bibinfo{journal}{IEEE Trans Robot}
  \bibinfo{year}{2015};31\bibinfo{volume}{(4)}:\bibinfo{pages}{850--863}.
\newblock \DOIprefix\doi{10.1109/TRO.2015.2430053}.
\bibitem[{Reyes-Uquillas and Hsiao(2021)}]{ref35}
\bibinfo{author}{Reyes-Uquillas D}, \bibinfo{author}{Hsiao T}.
  \bibinfo{title}{Safe and intuitive manual guidance of a robot manipulator
  using adaptive admittance control towards robot agility}.
\newblock \bibinfo{journal}{Robot Comput-Integr Manuf}
  \bibinfo{year}{2021}\bibinfo{volume}{;70}:\bibinfo{pages}{102127}.
\newblock \DOIprefix\doi{10.1016/j.rcim.2021.102127}.
\bibitem[{Chotiprayanakul et~al.(2007)Chotiprayanakul, Liu, Wang and
  Dissanayake}]{ref26}
\bibinfo{author}{Chotiprayanakul P}, \bibinfo{author}{Liu DK},
  \bibinfo{author}{Wang D}, \bibinfo{author}{Dissanayake G}. \bibinfo{title}{A
  3-dimensional force field method for robot collision avoidance in complex
  environments}.  In: \bibinfo{booktitle}{Proceedings of the 24th International
  Symposium on Automation and Robotics in Construction}; \bibinfo{year}{2007},
  p. \bibinfo{pages}{139--145}.
\bibitem[{Keemink et~al.(2018)Keemink, van~der Kooij and
  Stienen}]{keemink2018admittance}
\bibinfo{author}{Keemink AQ}, \bibinfo{author}{van~der Kooij H},
  \bibinfo{author}{Stienen AH}. \bibinfo{title}{Admittance control for physical
  human–robot interaction}.
\newblock \bibinfo{journal}{Int J Robot Res}
  \bibinfo{year}{2018};37\bibinfo{volume}{(11)}:\bibinfo{pages}{1421--1444}.
\newblock \DOIprefix\doi{10.1177/0278364918768950}.
\bibitem[{Chen and Ro(2022)}]{chen2022human}
\bibinfo{author}{Chen J}, \bibinfo{author}{Ro PI}. \bibinfo{title}{Human
  intention-oriented variable admittance control with power envelope regulation
  in physical human-robot interaction}.
\newblock \bibinfo{journal}{Mechatronics}
  \bibinfo{year}{2022}\bibinfo{volume}{;84}:\bibinfo{pages}{102802}.
\newblock \DOIprefix\doi{10.1016/j.mechatronics.2022.102802}.

\end{thebibliography}

\end{document}